\documentclass[letterpaper, 10 pt, conference]{ieeeconf}  

  \IEEEoverridecommandlockouts                              

\usepackage{etoolbox}
\usepackage{amsmath}
\usepackage{amssymb}
\usepackage{comment}
\usepackage{caption}
\usepackage{subcaption}
\usepackage{color}
\usepackage{textcomp}
\usepackage{paralist}
\usepackage{graphicx}      
\usepackage{cite}
\usepackage{url}
\usepackage{array}
\usepackage{makecell}
\usepackage[export]{adjustbox}
\usepackage{ntheorem}

\usepackage{mathtools}
\usepackage[utf8x]{inputenc} 
\usepackage[T1]{fontenc} 
\usepackage{bm} 

\usepackage{tikz}
\usetikzlibrary{matrix} 
\usetikzlibrary{calc} 
\DeclareMathOperator{\sgn}{sgn}

\tikzstyle{block} = [draw,rectangle,thick,minimum height=2em,minimum width=2em]
\tikzstyle{sum} = [draw,circle,inner sep=0mm,minimum size=2mm]
\tikzstyle{line} = [thick]
\tikzstyle{branch} = [circle,inner sep=0pt,minimum size=1mm,fill=black,draw=black]
\tikzstyle{guide} = []
\usepackage{fontawesome} 
\usetikzlibrary{positioning}
 
\renewcommand{\vec}[1]{\ensuremath{\boldsymbol{#1}}} 

\usetikzlibrary{graphs,graphs.standard,quotes}
\usetikzlibrary{arrows,decorations.pathmorphing,backgrounds,positioning,fit,petri}
\usetikzlibrary{shapes,arrows}
\usepackage{verbatim}

\newtheorem{theorem}{\textbf{Theorem}}

\newtheorem{myassump}{\textbf{Assumption}}
\newtheorem{myprob}{\textbf{Problem}}

\newtheorem{mylemma}{\textbf{Lemma}}

\newcommand\norm[1]{\left\lVert#1\right\rVert}
\title{\LARGE \bf
Leaderless Swarm Formation Control: From Global Specifications to Local Control Laws
}
\author{Solomon Gudeta, Ali Karimoddini, Mohammadreza Davoodi, and Ioannis Raptis  
\thanks{Solomon Gudeta, Ali Karimoddini, and Ioannis Raptis are with the Department of Electrical and Computer Engineering,
North Carolina Agricultural and Technical State University, Greensboro, North Carolina, USA.}
\thanks{Mohammadreza Davoodi is with the UT Arlington Research Institute, University of Texas at Arlington, Fort Worth, Texas, USA.}
\thanks{Corresponding author: A. Karimoddini, Tel: +13362853313, {\tt\small akarimod@ncat.edu}.}
}

\begin{document}
\maketitle
\thispagestyle{empty}
\pagestyle{empty}

\begin{abstract}
This paper introduces a distributed leaderless swarm formation control framework to address the problem of collectively driving a swarm of robots to track a time-varying formation. The swarm's formation is captured by the trajectory of an abstract shape that circumscribes the convex hull of robots' positions and is independent of the number of robots and their ordering in the swarm. For each robot in the swarm, given global specifications in terms of the trajectory of the abstract shape parameters, the proposed framework synthesizes a control law that steers the swarm to track the desired formation using the information available at the robot's local neighbors. For this purpose, we generate a suitable local reference trajectory that the robot controller tracks by solving the input-output linearization problem. Here, we select the swarm output to be the parameters of the abstract shape. For this purpose, we design a dynamic average consensus estimator to estimate the abstract shape parameters. The abstract shape parameters are used as the swarm state feedback to generate a suitable robot trajectory. We demonstrate the effectiveness and robustness of the proposed control framework by providing the simulation of coordinated collective navigation of a group of car-like robots in the presence of robots and communication link failures. 

\end{abstract}

\section{Introduction}
The use of robotic swarms in applications that are too risky for humans or where fast response is crucial and are beyond the capabilities of a single or few individual robots has recently received significant attention \cite{brambilla2013swarm}.
Tasks are assigned to the swarm in terms of reaching goals\cite{rubenstein2014programmable,bandyopadhyay2017probabilistic,miao2019multi,teruel2019distributed} and/or tracking a specified trajectory that capture the collective behaviors of the swarm\cite{morgan2016swarm,olfati2006flocking,belta2004abstraction,zhao2018affine}. In general, a swarm control problem involves the design of individual (local) robot controllers so that the swarm performs a specified collective (global) behavior required to execute a given task effectively.

There are various control design approaches presented in the literature of swarm control algorithms (see, e.g., \cite{soni2018formation,brambilla2013swarm} and reference therein), which include density-based  \cite{bandyopadhyay2017probabilistic}, potential field  \cite{olfati2006flocking,vanualailai2019stable}, optimization \cite{morgan2016swarm}, behavior-based \cite{weisbin2000nasa,lawton2003decentralized,rubenstein2014programmable}, consensus-based  \cite{joordens2010consensus,liu2018decentralization,hong2006tracking}, leader-follower\cite{KARIMODDINI2013424,6858770, loria2015leader,KARIMODDINI2011886}, and virtual structure control\cite{liu2018decentralization} methods, to name a few. 
 The density-based, potential field-based and optimization approaches are mainly employed to solve the swarm pattern formation problem \cite{bandyopadhyay2017probabilistic,vanualailai2019stable,savkin2016distributed,morgan2016swarm}. 
Similarly, behavior-based approaches have been employed to design local behaviors for robots to perform swarm navigation to achieve the desired performance  collectively. The work in \cite{lawton2003decentralized} develops a decentralized behavior-based architecture,  requiring fewer communications among the robots in the swarm. However, in general, behavior-based approaches are analytically challenging to establish proofs of their convergence.
On the other hand, in the leader-follower and virtual structure methods, the desired trajectory of the swarm are assigned to leader robot(s),  virtual leader(s), or virtual structures.  Olfati-saber \cite{olfati2006flocking} employed formation graphs to capture the robots' dynamics and inter-robot constraints, and then combined them with a potential field and virtual leader approach to drive a group of agents along a specified path. 
To improve the scalability of the swarm control algorithm, Belta et al. \cite{belta2004abstraction} proposed an abstraction-based control framework that drives a swarm of robots along a given path. However, the centralized architecture in \cite{belta2004abstraction} makes the design vulnerable to observer failures and communication link losses. 
 Recently, Shiyu Zhao \cite{zhao2018affine} presented a new approach based on stress matrices of graphs to achieve multi-agent formation maneuvers. The author adopted a distributed leader-follower approach to solve the formation maneuver control problem for a team of single-integrator, double-integrator, unicycle, and non-holonomic agents. However, the calculation stress matrices in \cite{zhao2018affine} is nontrivial. 
Freeman et al. \cite{freeman2006distributed} designed a distributed estimation algorithm to estimate first-order and second-order moments of the swarm's distribution. They combined their estimation algorithm with motion controllers for each robot to regulate the shape and position of the swarm  \cite{yang2008multi}. Nevertheless, the proposed PI estimator exhibits slow convergence rates, and the combined estimator/controller algorithm in \cite{freeman2006distributed} is limited to single integrator agents. 

In this paper, given a global specification (swarm formation and trajectory), we propose a scalable and robust distributed control framework for synthesizing control laws for local (individual) robots so that they, as a group, can switch to any time-varying affine transformation of initial swarm formation while the swarm is tracking a desired bounded $C^1$ trajectory.
For this purpose, we develop a distributed control algorithm for swarm formation control using feedback linearization and dynamic average consensus estimation. 
A salient feature of the proposed method is handling the losses or addition of robots from/to the swarm. This is due to the flexible and distributed architecture of the proposed framework versus the fixed and centralized network architecture in \cite{belta2004abstraction} where a communication loss with an observer results in a complete failure of the swarm system.
Furthermore, many existing leader-follower based swarm algorithms require robot labeling \cite{loria2015leader, zhao2018affine}. For example, in \cite{loria2015leader}, a unique swarm leader and a unique tail robot are required for the swarm to navigate along the desired trajectory. However, our leaderless swarm formation control formulation do not require special robot ordering and labeling.
More importantly, compared to swarm algorithms with a fixed inter-robot distance (see, e.g. \cite{olfati2006flocking}), our control design formulation allows the swarm to shrink, expand, rotate, translate, or perform compositions of these operations. All these features make our swarm control framework suitable for diverse applications.

The organization of the rest of this paper is as follows. In Section \ref{sec:form_and_tra_prob}, the leaderless swarm formation control problem is formulated. In Section \ref{sec:form_and_tra_framework}, we propose a distributed control law 
for a swarm of robots to track the desired time-varying formation.
In Section \ref{sec:form_and_tra_simu}, we provide simulation results to verify the effectiveness of the developed control framework. Finally, concluding remarks are synopsized in Section \ref{sec:form_and_tra_conc}. 

\section{Problem formulation}
\label{sec:form_and_tra_prob}
Consider a swarm $\mathcal{S}$ of $N$ identical rear-wheel driving car-like robots deployed to execute task $\mathcal{T}$ in a world-frame $\mathcal{W}$ ( with center $O_{\mathcal{W}}$ and basis vectors
 $\{\vec{x}_{\mathcal{W}},\vec{y}_{\mathcal{W}}\}$).
The governing kinematics of Robot $i$ are given by 
\begin{equation}
\dot{x}_i = g_i(x_i)u_i, \quad i=1,\cdots,N, \\
\label{eq:robotM}
\end{equation}
where $x_i = \begin{bmatrix}\bar{x}_i& \bar{y}_i&\theta_i&\phi_i \end{bmatrix}^\mathrm{T}\subseteq \mathbb{R}^4$ is the state vector, $u_i = [ v_i \quad \omega_i]^{\mathrm{T}} \subseteq \mathbb{R}^2$ is the control input vector, 
$[\bar{x}_i \quad \bar{y_i}]^{\mathrm{T}} $ is the position vector, $\theta_i$ is the heading angle, $\phi_i$ is the steering angle, $v_i $ is the linear velocity, $\omega_i$ is the steering velocity, and  $g_i(x_i) = \begin{bmatrix} g_{i_1}(x_i) & g_{i_2}(x_i)\end{bmatrix}$, where $g_{i_1}(x_i) = \begin{bmatrix} \cos \theta_i & \sin \theta_i & \frac{1}{L}\tan \phi_i & 0 \end{bmatrix}^\mathrm{T}$, $g_{i_2}(x_i)= \begin{bmatrix} 0 & 0 & 0 & 1 \end{bmatrix}^\mathrm{T}$, and $L$ is the wheel base of Robot $i$, respectively. Let $\mathcal{G}(t) = (\mathcal{V}(t),\mathcal{E}(t))$ be a time-varying communication graph of the swarm $\mathcal{S}$ at time $t$, where $\mathcal{V}(t)$ is the set of robots in the swarm  and $\mathcal{E}(t) \subseteq \{(i,j): i, j \in \mathcal{V}(t), i \neq j\}$ is the set of communication links among the robots in the swarm. The communication graph $\mathcal{G}(t)$ changes i) when new robots join the swarm; ii) when the swarm loses some member robots, and iii) when the communication links among the member robots fail. 
\begin{myassump}
The communication graph $\mathcal{G}(t)$ is assumed to be a slowly time-varying graph. Also, we assume that  $\mathcal{G}(t)$ is a strongly connected bidirectional graph at each time $t$. 
\end{myassump}
 Let the set of neighbors of Robot $i$ at time $t$ be given by $\mathcal{N}_i(t) = \{j \in \mathcal{V}(t): (i,j) \in \mathcal{E}(t)\}$.
 The swarm configuration $x_s \in X_s$ of a swarm $\mathcal{S}$ is defined as $x_s =  col_{i=1}^N(x_{s_i})$,
 where $x_{s_i} =[\bar{x}_{i} \quad \bar{y}_i]^\mathrm{T} $, $i = 1,\cdots,N$, and the operator $col(\cdot)$ stacks the argument vectors.
 The swarm structure $\chi$  is then defined as the tuple $\chi = (\mathcal{V}(t), \mathcal{E}(t), x_s)$.
Now, our objective is to design a swarm formation control law $u$ that steers
a swarm of robots $\mathcal{S}$ given by
\begin{equation}
\dot{x} = G(x)u,
\label{eq:swarmsys}
\end{equation}
where $x = col_{i=1}^N(x_i)$, $G(x)$ = diag$(g_1(x_1),\cdots,g_N(x_N))$, $u = col_{i=1}^N(u_i) $,
along the desired path. Given a large number of robots evolving in the swarm configuration space $X_s$, solving the aforementioned control problem is non-trivial as the dimension of the swarm system in (\ref{eq:swarmsys}) depends on the number of robots in the swarm. To remedy this, we capture the motion of the swarm 
in terms of the motion of an abstract shape.
An abstract shape is a convex closed curve $\mathbb{S}$ circumscribing the convex hull of  configuration $x_s$ of the swarm structure $\chi$. In the Cartesian coordinate ($\bar{x}$, $\bar{y}$), the abstract shape $\mathbb{S}$ is given by
\begin{equation}
\big\lvert\frac{\bar{x}-\mu_x}{s_w}\big\rvert^{m_a}+\big\lvert\frac{\bar{y}-\mu_y}{s_l}\big\rvert^{n_a} = 1, n_a \geq 2, m_a \geq 2,
\end{equation}
where $(\mu_x, \mu_y)$ is the coordinate of center of the abstract shape, $s_l$ and $s_w$ are the length of the semi-major and semi-minor axes of the abstract shape, respectively.
Let $a \in \mathbb{R}^5$ be the vector of parameters of abstract shape  (see Section \ref{sec:form_and_tra_framework}). Also, let a surjective submersion
\begin{equation}
\Phi: \mathbb{R}^{2N} \rightarrow \mathbb{R}^5, \quad  \Phi(x_s) = a.
\label{eq:abstraction_map}
\end{equation}
relate the abstract shape parameters $a$ and the swarm configuration $x_s$.
 Then, we can specify the desired path for the swarm as the trajectory of the abstract shape  parameters $a$. Specifying the swarm's desired trajectory in terms of the trajectory of the abstract shape  parameters is more practical than providing the desired trajectory for every member robot in the swarm. Also, in this approach, the swarm's trajectory is independent of the number and permutation of robots in the swarm.

Now, we aim to design a distributed control law $u_i$ for each Robot $i$ in the swarm so that the abstract shape  parameters $a$ track the desired trajectory $\zeta$ specified by the user (motion planner or human). However, the synthesis of control law $u_i$ requires each Robot $i$ in the swarm to know the position of all robots in the swarm to determine the abstract shape parameters $a$. In \cite{belta2004abstraction},  this requirement is handled by introducing a central observer that moves with the swarm. The observer collects the position information of all robots in the swarm, computes the abstract shape parameters, and broadcasts the computed value of the abstract shape parameters to all robots in the swarm. 
However, this approach requires all the robots to be in the communication range of each other or the observer, thus, prone to high bandwidth requirements or a single point of failure at the observer.
To circumvent this problem and realize a distributed control architecture, we design a dynamic average consensus estimator for each Robot $i$ to estimate the abstract shape parameters $a$ only based on the information collected from Robot $i$'s neighboring robots.
More precisely, we state the formation and trajectory tracking control problem as follows:
\begin{myprob} Under Assumption 1, given a time varying  desired trajectory $\zeta \in \mathbb{R}^5$ of an abstract shape,
\vspace{-0.1cm}
\begin{enumerate}[a)]
\item Construct the abstract shape parameters $a$.
\item For the abstract shape parameters $a$, design a distributed neighbor-based estimator so that each robot's estimation of abstract shape parameters $\bar{a}_i$ converges to $a$ in a finite time,
for all $i \in \{1,\cdots,N\}$.
\item Generate the desired trajectory for Robot $i$, $i \in \{1,\cdots,N\}$, such that the abstract shape parameters $a$ tracks $\zeta$.
\item Design a feedback control law $u_i$ for Robot $i$, $i \in \{1,\cdots,N\}$, so that each robot tracks its desired trajectory, and the swarm system (\ref{eq:swarmsys}) tracks the given trajectory $\zeta \in \mathbb{R}^5$.
\end{enumerate} 
\label{prob:prob2-decs}
\end{myprob}
   
\section{The leaderless swarm formation control framework }
\label{sec:form_and_tra_framework}
A swarm formation control is a challenging problem. Complexities of the robot kinematics and the swarm dynamics often lead to intractable control problems. This section employs tools from differential geometry, consensus, and control theory to systematically design local control laws for Robot $i$ in the swarm to realize a specified swarm formation $\zeta$. The desired time-varying swarm formation is given in terms of the abstract shape parameters $a$. For Robot $i$, we propose a control law that drives  
the pose and shape of the swarm (the motion of the abstract shape) to track the desired trajectory $\zeta$. For this purpose, Robot $i$ estimates the value of the abstract shape parameters $a$ via a dynamic consensus estimator
from the information available at neighboring robots. The detailed design of the proposed control framework is presented in the following sections.
\subsection{Trajectory-tracking control law}
This Section solves a trajectory tracking control problem (Problem \ref{prob:prob2-decs}.d) using input-output linearization. For this purpose, we define the Robot $i$'s output $ y_i=h_i(x_i)$, $h_i(x_i) :\mathbb{R}^4\rightarrow \mathbb{R}^\mathfrak{r}$ in such a way that the decoupling between Robot $i$'s linear input-output dynamics and internal dynamics is achieved:
\begin{equation}
\begin{split}
h_{i_1} &= x_{i_1}+L\cos x_{i_3}+D\cos(x_{i_3}+x_{i_4})\\
h_{i_2} &= x_{i_2}+L\sin x_{i_3}+D\sin(x_{i_3}+x_{i_4}),
\end{split}
\end{equation}
where $\mathfrak{r}$ is the total relative degree of Robot $i$ and $D \neq 0$ is the ``look-ahead'' distance.
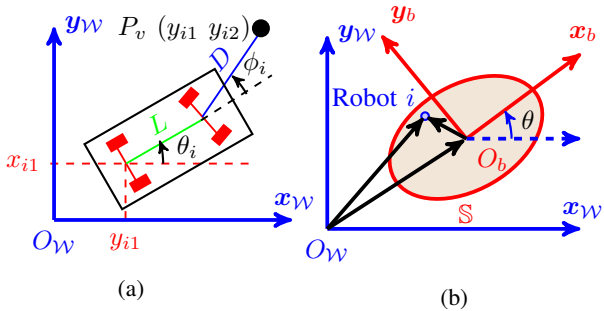
\begin{figure}[b]
\begin{subfigure}{0.2\textwidth}
\centering
\begin{tikzpicture}[-,>=stealth',shorten >=1pt,auto,node distance=2cm,
                    thick,main node/.style={circle,fill=blue!20,draw,font=\sffamily\Large\bfseries}]
 \draw [line width=1.5mm, red, rotate=30 ] (-2.4,-1.2) -- (-2.1,-1.2);
 \draw [line width=1.5mm, red, rotate=30 ] (-2.4,-1.82) -- (-2.1,-1.82);
 \draw [line width=1.5mm, red, rotate=50 ] (-1.6,-0.7) -- (-1.3,-0.7);
 \draw [line width=1.5mm, red, rotate=50 ] (-1.7,-1.35) -- (-1.4,-1.35);
 \draw [line width=0.25mm, red,rotate=30] (-2.3,-1.2) -- (-2.3,-1.82);
 \draw [line width=0.25mm, red,rotate=50] (-1.5,-0.7) -- (-1.6,-1.35);
 \draw [line width=0.25mm, green,rotate=30] (-2.3,-1.5) -- node [anchor = center,above, rotate = 30, midway] {$L$} (-1.05,-1.5);
 \draw [line width=0.25mm, blue,rotate=50] (-1.55,-1.0) -- node [anchor = center,above, rotate = 50, midway] {$D$} (-0.05,-0.85);
 \draw [dashed,line width=0.25mm, black,rotate=30] (-1.15,-1.5) -- (-0.0,-1.45);
 \draw [dashed,line width=0.25mm, red,rotate=0] (-2.25,-2.45) node [anchor = east] {$x_{i1}$} -- (0.5,-2.45);
 \draw [dashed, line width=0.25mm, red,rotate=0] (-1.25,-2.45) -- (-1.25,-3.25)node [anchor = north] {$y_{i1}$};
 \filldraw (0.55,-0.65) circle (3pt) node [anchor = east] {$P_v$ $(y_{i1}$ $ y_{i2})$};
  \draw[->] (-0.75,-2.45) arc (0:20:1) node at (-0.7,-1.95) [anchor = north west] {$\theta_{i}$};
   \draw [->,rotate=30](-0.45,-1.5) arc (0:20:1) node at (-0.4,-1.06) [anchor = west] {$\phi_{i}$};
\draw[draw=black, rotate=30] (-2.65cm,-2.0cm) rectangle ++(2.0,1.0);
\draw [line width=0.5mm, blue, ->](-2.2,-3.2)node [anchor=north] {$O_\mathcal{W}$} -- (-2.2,-0.6) node [anchor=west] {$\vec{y}_{\mathcal{W}}$};
\draw [line width=0.5mm, blue, ->](-2.2,-3.2) -- (1.0,-3.2) node [anchor=south] {$\vec{x}_{\mathcal{W}}$};      
\end{tikzpicture}
\caption{}
\label{fig:virtualpt}
\end{subfigure}
\begin{subfigure}{0.27\textwidth}
\centering
\begin{tikzpicture}[->,>=stealth',shorten >=1pt,auto,node distance=2cm,
                    thick,main node/.style={circle,fill=blue!20,draw,font=\sffamily\Large\bfseries}]
\draw [rotate=30] [line width=0.5mm,,red,fill=brown!20](-1.3,-1.5392) ellipse (1.1cm and 0.7cm) node [anchor=north, below = 0.75cm] {$\mathbb{S}$};
\draw [blue,fill=blue!20] (-0.9,-1.7) circle (0.05cm) node [anchor= south east] {Robot $i$};
\draw [line width=0.5mm, red](-0.35,-2.0)node [anchor=north west] {$O_b$} -- (-1.5,-0.6) node [anchor= south west] {$\vec{y}_b$};
\draw [line width=0.5mm, red](-0.35,-2.0) -- (1.2,-0.85) node [anchor=south] {$\vec{x}_b$};
\draw [line width=0.5mm, blue](-2.2,-3.2)node [anchor=north] {$O_\mathcal{W}$} -- (-2.2,-0.6) node [anchor=west] {$\vec{y}_{\mathcal{W}}$};
\draw [line width=0.5mm, blue](-2.2,-3.2) -- (1.2,-3.2) node [anchor=south] {$\vec{x}_{\mathcal{W}}$};
\draw [dashed,line width=0.5mm, blue](-0.35,-2.0) -- (1.2,-2.0);
\filldraw[fill=cyan, draw=blue] (0.25,-2.0)  node [anchor =south west] {$\theta$} arc (0:30:0.8) ;
\draw [line width=0.5mm, black](-2.2,-3.2) -- (-0.9,-1.7);   
\draw [line width=0.5mm, black](-2.2,-3.2) -- (-0.35,-2.0);  
\draw [line width=0.5mm, black](-0.35,-2.0) -- (-0.9,-1.7); 
\end{tikzpicture}
\caption{}
\label{fig:abstractshape}
\end{subfigure}
\caption{(a) Robot reference frames and definition of virtual point $P_v$, (b) Swarm reference frames and the abstract shape  that circumscribes the region occupied by swarm of robots.}
\end{figure}
The output function $h_i$, defines the position of a virtual point $P_v$ (see Figure \ref{fig:virtualpt}) in front or behind of Robot $i$ based on the sign of $D$ to simplify the control design  by decoupling input-output dynamics and internal dynamics.
Let the augmented function $\bar{h}_i(x_i): \mathbb{R}^4 \rightarrow \mathbb{R}^{4-\mathfrak{r}}$ be chosen as $\bar{h}_i(x_i) = \begin{bmatrix} x_{i_3} \quad x_{i_4} \end{bmatrix}^\mathrm{T}$ such that the state transformation $T_i(x_i)=
\begin{bmatrix} q_i^\mathrm{T} \quad \check{q}_i^\mathrm{T}\end{bmatrix}^\mathrm{T}  = \begin{bmatrix} h_i^\mathrm{T}(x_i) \quad \bar{h}_i^\mathrm{T}(x_i)\end{bmatrix}^\mathrm{T}$ is a diffeomorphism. 
Also, let the control input $u_i$ to Robot $i$ be given as $u_i = \bar{\alpha}_i(x_i)+\bar{\beta}_i(x_i)\bar{v}_i $, where $
\bar{\alpha}_i(x_i) = 0$ due to the kinematic model being drift free, and $
\bar{\beta}_i(x_i) = \Delta^{-1}_i(x_i) $, where $
\Delta_i(x) = 
\begin{bmatrix}
L_{g_{i_1}}h_{i_1}(x_i) & L_{g_{i_2}}h_{i_1}(x_i)\\
L_{g_{i_1}}h_{i_2}(x_i) & L_{g_{i_2}}h_{i_2}(x_i)
\end{bmatrix}$ is the decoupling matrix,  $L_{g_{i_j}}h_{i_k}(x_i)$ is the Lie derivative of function $h_{i_k}(x_i)$ along a vector field $g_{i_j}(x_i)$, for $j \in \{1,2\}$ and $k \in \{1,2\}$.
Using the state transformation $T_i(x_i)$ and state feedback control law $u_i$, we transform (\ref{eq:robotM}) into 
\begin{equation}
\dot{\check{q}}_i = \check{f}_i(q_i,\check{q}_i),
\dot{q_i} = A_iq_i+B_i\bar{v}_i,
y_i = C_iq_i,
\label{eq:robotM2}
\end{equation}
\noindent where $q_i = [h_{i_1} \quad h_{i_2}]^{\mathrm{T}}$, $\dot{\check{q}}_i = \check{f}_i(q_i,\check{q}_i)= \frac{\partial \bar{h}_i(x_i)}{\partial x_i}\dot{x}_i$ is the internal dynamics, and $
A_i = \left[\begin{smallmatrix} 0&0\\
                      0&0\\
                      \end{smallmatrix}\right]$, $
B_i = I_{2}$, $C_i = I_{2}$, $ \bar{v}_i = \begin{bmatrix} \bar{v}_{i_1}\quad \bar{v}_{i_2}\end{bmatrix}^\mathrm{T}$.
Based on this, we design a linear control law $\bar{v}_i$ so that $q_i$ can track desired trajectory ${q_{i}}_d$.  Assuming that all states of the control affine system in (\ref{eq:robotM}) are measurable,  for stable zero dynamics \cite{wang2003full}, we employ the control law
\begin{equation}
\bar{v}_i = \dot{q}_{id}+\check{K}_{i}(q_{id} - q_{i}),\check{K}_{i} > 0,
\label{eq:robotctrl1}
\end{equation}
where $\check{K}_{i}$ is the control gain, for Robot $i$ to exponentially tracks the desired trajectory $q_{id}$.
In many application area of swarm of robots, specifying reference trajectory $q_{id}$ for each individual robot is not practical. Therefore, we design $q_{id}$ by solving the  multi-input  multi-output (MIMO) input-output linearization problem in the next section. 
\subsection{Trajectory generation}
The collective behavior of robots in the swarm is captured by the motion of the abstract shape, which is represented by the trajectory of the abstract shape parameters $a$. For each robot in the swarm, we design a distributed swarm controller so that the abstract shape parameters $a$ tracks the desired trajectory $\zeta$. In this setting, we design a dynamic average consensus estimator to estimate the abstract shape parameters $a$. Then, for each Robot $i$, the swarm controller output will be converted to the desired trajectory $q_{id}$ to be tracked by the trajectory tracking control law (\ref{eq:robotctrl1}). For Robot $i$, the approaches to the desired trajectory generation are discussed next.
We start by putting together the linearized input-output dynamics given in (\ref{eq:robotM2}) to form a new swarm system as
\begin{equation}
\dot{q} =  Aq+B\bar{v}, \quad  y= Cq,
\label{eq:robotM3}
\end{equation}
where $A$ =  diag$(A_1,\cdots,A_N)$, $B$ =  diag$(B_1,\cdots,B_N)$, $C$ =  diag$(C_1,\cdots,C_N)$, $q= col_{i=1}^N(q_i)$, $\bar{v}= col_{i=1}^N(\bar{v}_i)$.
The output $y$ of the swarm system in (\ref{eq:robotM3}) is the  collection of the output of individual robots (local behaviors). However, the control specifications for the swarm is given in terms of collective (global) behaviors of the swarm. To address this issue, we transform (\ref{eq:robotM3}) from the robot configuration space to the abstract shape space using the input-output linearization technique.
To input-output linearize the swarm system in (\ref{eq:robotM3}), we construct a new output function $\tilde{y}$ to be the abstract shape parameters $a$ of abstract shape.
Let the abstract shape be described in the world coordinate by Frame $\{\mathcal{W}\}$ (shown in Figure~\ref{fig:abstractshape} with center $O_{\mathcal{W}}$ and basis vectors
 $\{\vec{x}_{\mathcal{W}},\vec{y}_{\mathcal{W}}\}$) and in the body coordinate by Frame $\{b\}$  (shown in Figure~\ref{fig:abstractshape} with center $O_b$ and basis vectors $\{\vec{x}_b,\vec{y}_b\}$). 
 The position vector of virtual point of Robot $i$ with respect to Frame $\{\mathcal{W}\}$ is represented by $q_i$, and the position vector describing the origin of  Frame $\{b\}$ with respect to Frame $\{\mathcal{W}\}$ is denoted by $O_b^\mathcal{W}$. Let $R_b^\mathcal{W} \in SO(2)$ be the rotation matrix of Frame $\{b\}$ with respect to Frame $\{\mathcal{W}\}$ and $p_i$ be the position vector of virtual point of Robot $i$ with respect to Frame $\{b\}$. Using geometry, $p_i$ is given by
\begin{equation}
p_i = \begin{bmatrix}p_{ix} & p_{iy}\end{bmatrix}^\mathrm{T}= -{R_b^\mathcal{W}}^{\mathrm{T}}O_b^\mathcal{W}+{R_b^\mathcal{W}}^{\mathrm{T}}q_i,
\label{eq:body:refframe}
\end{equation} 
where $p_{ix}$ and $p_{iy}$ are the components of vector $p_i$ in Frame $\{b\}$.
The origin $O_b^\mathcal{W}$ of Frame $\{b\}$, $\mu$, is the mean of position vectors of virtual point of each Robot $i$ in Frame $\{\mathcal{W}\}$. It represents the center of the abstract shape :
$
O_b^\mathcal{W} = \mu = \frac{1}{N}\sum_{i=1}^Nq_i.
$
We utilize the co-variance matrix of robot distribution in the region circumscribed by the abstract shape to define the shape and orientation of the swarm.  The co-variance matrix of the ensemble of the robots in Frame $\{b\}$ is given by

\small
\begin{equation}
\Sigma_1 = 
\begin{bmatrix}
\frac{1}{N-1}\sum_{i=1}^N(p_{ix}-0)^2 & 0 \\
0 & \frac{1}{N-1}\sum_{i=1}^N(p_{iy}-0)^2
\end{bmatrix}.
\label{eq:sigma1}
\end{equation}
\normalsize
Similarly, the co-variance matrix of the ensemble of the robots in Frame $\{\mathcal{W}\}$ is given by
\begin{equation}
\begin{split}
\Sigma_0 = 
\begin{bmatrix}
\sigma_{xx} & \sigma_{xy}\\
\sigma_{xy} & \sigma_{yy}
\end{bmatrix}.
\end{split}
\end{equation}
The covariance matrix $\Sigma_0$ is related to the co-variance matrix $\Sigma_1$ in Frame $\{b\}$ as
\begin{equation}
\Sigma_1 = {R_b^\mathcal{W}}^\mathrm{T}\Sigma_0 R_b^\mathcal{W} = \begin{bmatrix}
s_1 & s_{12}\\
s_{12} & s_2
\end{bmatrix}.
\label{eq:body:cov1}
\end{equation}
 Solving (\ref{eq:body:cov1}), the orientation $\theta$ of the abstract shape will be
\begin{equation}
\theta = \frac{1}{2}tan^{-1}\big(\frac{2\sigma_{xy}}{\sigma_{yy}-\sigma_{xx}}\big).
\label{eq:theta}
\end{equation}

Now, consider the convex hull that captures $p$ percentage of robots in the swarm. Then, the width and length of the abstract shape  in Frame $\{b\}$ can  be captured by $s_w = \sqrt{c_ps_2}$ and $s_l = \sqrt{c_ps_1}$, respectively, where $c_p = -2\ln(1-p)$. 
The abstract shape  parameters is given by a $5-$dimensional vector
$a = \begin{bmatrix}
\mu=[\mu_x,\, \mu_y],\,\theta,\,s_2,\, s_1
\end{bmatrix}^\mathrm{T}$. 
Besides, assuming that $q \approx x_s$, from Definition \ref{eq:abstraction_map} we have  $\Phi(x_s)\approx \Phi(q) = a$. 
 Now, to address Problem \ref{prob:prob2-decs}.a, we define the mapping $\Phi$ as $\Phi(q)=a = [\Phi_1(q) \cdots \Phi_5(q)]^\mathrm{T}$, where $\Phi_1(q) = \mu_x$, $\Phi_2(q) = \mu_y$, $\Phi_3(q) = \theta$, $\Phi_4(q) = s_2$, and $\Phi_5(q)= s_1$. Then, the state feedback control law will be 
\begin{equation} 
\bar{v} = \alpha(q)+\beta(q)w,
\label{eq:swarm_controller_fb}
\end{equation}
\noindent where
\begin{equation*}
\begin{split}
\alpha(q) &= -\Delta_s^{-1}(q).\begin{bmatrix}L_{f}^{r_{i_1}}\Phi_1(q) & \cdots & L_{f}^{r_{i_5}}\Phi_5(q) \end{bmatrix}^\mathrm{T}=0,\\
\Delta_s(q) &= \begin{bmatrix}
L_{g_1}L_{f}^{r_{i_1}-1}\Phi_1(q) & \cdots & L_{g_N}L_{f}^{r_{i_1}-1}\Phi_1(q)\\
\vdots & \ddots &\vdots\\
L_{g_1}L_{f}^{r_{i_5}-1}\Phi_5(q) & \cdots & L_{g_N}L_{f}^{r_{i_5}-1}\Phi_5(q)\\
\end{bmatrix},\\ 
\Delta_s(q) &= \begin{bmatrix}
\frac{I_2}{N} & \cdots & \frac{I_2}{N}\\
\vdots & \ddots &\vdots\\
\frac{(q_1-\mu)^\mathrm{T}R_{s_1}}{N-1} &\cdots& \frac{(q_N-\mu)^\mathrm{T}R_{s_1}}{N-1}\\
\end{bmatrix}, \begin{matrix}\beta(q) = \Delta_s^{-1}\\f= Aq=0\end{matrix},\\
R_{s_1}&= \begin{bmatrix}2\cos^2{\theta} & \sin{2\theta}\\ \sin{2\theta} &2\sin^2{\theta}\end{bmatrix}, 
\quad  \begin{bmatrix}g_1 & \cdots & g_N\end{bmatrix}= B, \quad \\
\end{split}
\end{equation*}
where $r_{i_j} \in \mathbb{N}$, $i = {1,\cdots,N}$, $j= {1,\cdots,5}$, is the vector relative degree of (\ref{eq:robotM3}), $L_{f}\Phi_j(q)$ is the Lie derivative of function $\Phi_j(q)$ along a vector field $f$, and $L_{g_k}L_{f_i}\Phi_j(q)$ is the Lie derivative of function $\Phi_j(q)$ along a vector field $f_i$ and along another vector field $g_k$, where $k = {1,\cdots,N}$. The surjective submersion $\Phi(q)$ and the state feedback control law $\bar{v}$ in (\ref{eq:swarm_controller_fb}) transforms (\ref{eq:robotM3}) into 
\begin{equation}
\Xi_2 = 
\begin{cases}
\dot{a}(t) = \bar{A}a+\Bar{B}w,\quad \quad
\tilde{y} = a,
\end{cases}
\label{eq:robotM4}
\end{equation}
where $\bar{A} = 0_{5\times5}$,  $\bar{B} = I_{5\times5}$, and
$w$ is the control law that steers abstract shape parameter $a$ to track the desired trajectory $\zeta$.
For the sake of reducing the computation cost, we simply design  $w$ as a linear control law, given as
\begin{equation}
w = \bar{K}(\zeta-\bar{a})+\dot{\zeta},
\label{eq:robotTr1}
\end{equation}
where $\bar{K}$ is the control gain and $\bar{a}$ is the estimate of $a$. 

Now, we can calculate $\bar{v}_i $ from $w$ as $\bar{v}_i = \alpha_i(q_i)+\beta_i(q_i)w_i$, where $\alpha_i(q_i)$ and $\beta_i(q_i)$ are $i^{th}$ row of $\alpha(q)$ and $\beta(q)$, respectively. We then compute the reference trajectory $q_{id}$ by solving
\begin{equation}
\dot{q}_{id}  = \alpha_i(q_i)+\beta_i(q_i)(\bar{K}(\zeta-\bar{a})+\dot{\zeta}), 
\label{eq:robotTr}
\end{equation}
with $q_i(0) = x_{s_i}(0)$ being known. 
\subsection{The dynamic  consensus estimator}
Determining the abstract shape parameters vector $a\in R^5$ requires a centralized communication architecture or all$-$to$-$all communication among the robots in the swarm. This process is prone to failures associated with the centralized observer and with communication links between the observer and individual robots in the swarm. Therefore, rather than relying on a central observer to compute the abstract shape state vector $a$,  we estimate the abstract shape state vector $a$ (Problem \ref{prob:prob2-decs}.b) by exploiting the underlying graph structure of the network of robots in a distributed way using dynamic average consensus.  
To leverage the technique of average consensus, we represent all the components of the abstract shape parameters $a$ in terms of the average of suitable  expressions. First, we re-write (\ref{eq:theta}) as $\theta = \frac{1}{2}tan^{-1}\big(\frac{\sigma_1^*}{\sigma_2^*}\big)$, where $\sigma_1^* = \frac{2\sigma_{xy}}{N}
$ and $
\sigma_2^* = \frac{\sigma_{yy}-\sigma_{xx}}{N} $. Then, we introduce $z_i$ as:
\begin{equation}
z_i = \begin{bmatrix}z_{i_1}\\ z_{i_2}\\z_{i_3}\\z_{i_4}\\z_{i_5}\\z_{i_6}\end{bmatrix} = \begin{bmatrix} q_{ix} \\
q_{iy}\\
 2(q_{ix}-\mu_x)(q_{iy}-\mu_y)\\
(q_{iy}-\mu_y)^2-(q_{ix}-\mu_x)^2\\
 p_{ix}^2 \\
 p_{iy}^2 
\end{bmatrix}.
\end{equation}
Accordingly, for swarm of $N$ robots, parameters $\left[\begin{smallmatrix}\mu_x & \mu_y & \sigma_1^* & \sigma_2^* & s_1 & s_2\end{smallmatrix}\right]^{\mathrm{T}}$ are expressed as the average of $z_i$, that is, $\frac{1}{N}\sum_{i = 1}^Nz_{i_k}$. To estimate $\frac{1}{N}\sum_{i = 1}^Nz_{i_k}$, where $k = {1,\cdots,6}$, we implement an edge-based dynamic consensus  estimator of the form
\begin{equation}
\begin{split}
\dot{\eta}_{ij_k}^+ &= -\rho\tanh\{c(\gamma_{i_k}-\gamma_{j_k})\}\\
\dot{\eta}_{ij_k}^- &=
-\rho\tanh\{c(\gamma_{j_k}-\gamma_{i_k})\}, c\geq 1, j\in \mathcal{N}_i\\
\gamma_{i_k} &= \sum_{j\in \mathcal{N}_i}{\eta}_{ij_k}^+-\sum_{j\in \mathcal{N}_i}{\eta}_{ij_k}^-+z_{i_k}, \quad  k = 1,\cdots,6,
\end{split}
\label{eq:consensus}
\end{equation}
\noindent where ${\eta}_{i} = [{\eta}_{ij}^+ \quad {\eta}_{ij}^-]^\mathrm{T} \in \mathbb{R}^{2\mathcal{N}_i}$ is the internal state of the estimator on Robot $i$, $\rho \in \mathbb{R}$ and $c \in \mathbb{R}$ are global estimator parameters, and  $\gamma_{i_k} \in \mathbb{R}$ is the estimate of  $\frac{1}{N}\sum_{i = 1}^Nz_{i_k}$  where $k = {1,\cdots,6}$. From (\ref{eq:consensus}), it is clear that the edge dynamics captures the state of the disagreement between Robot $i$ and Robot $j$.  Further, the use of $tanh(.)$ in (\ref{eq:consensus}), makes the proposed estimator smooth, avoiding the chattering phenomena  \cite{gudeta2020robust}. This approach makes the protocol robust to agents joining or leaving the network, and to communication link failures among the agents. The proposed estimator has three stages due to the fact that the estimation of the average of some of the components of $z_i$ requires the knowledge of the average of other components of $z_i$. In the first stage, we estimate the average of $z_{i_1}$ and $z_{i_2}$ by the mean estimator. Using the information from the mean estimator stage, the average of $z_{i_3}$ and $z_{i_4}$ is then estimated in the second stage by the orientation estimator. Similarly, in the third stage, using the information from the orientation estimator, we estimate the average of $z_{i_5}$ and $z_{i_6}$ by the width and length estimator. 
Then, the estimate  of the components of abstract shape parameters $a$ at each Robot $i$ is given by $ \bar{\mu} = \begin{bmatrix}\gamma_{i_1}& \gamma_{i_2}\end{bmatrix}$, $\bar{\sigma}_1^* = \gamma_{i_3}$, $\bar{\sigma}_2^*= \gamma_{i_4}$,  $\bar{s}_1 = \gamma_{i_5}$, and $\quad\bar{s}_2 = \gamma_{i_6}$. Based on this, the estimation of the abstract shape orientation is given as $\bar{\theta}= \frac{1}{2}tan^{-1}(\frac{\bar{\sigma}_1^*}{\bar{\sigma}_2^*})$. Further, the estimate of the length of semi-minor axis $\bar{s}_w$ and the semi-major axis $\bar{s}_l$ of the abstract shape are given as $\bar{s}_w = \sqrt{c_p\bar{s}_2}$ and $\bar{s}_l = \sqrt{c_p\bar{s}_1} $, respectively. Accordingly, the estimate of the abstract shape parameters $\bar{a}$
is given by 
\begin{equation}
\bar{a}(t) = \begin{bmatrix}
 \bar{\mu} & \bar{\theta} &\bar{s}_2 & \bar{s}_1
\end{bmatrix}^\mathrm{T}.
\label{eq:a_bar}
\end{equation}

\section{Simulation results}
\label{sec:form_and_tra_simu}
\begin{figure}
\vspace{0.25cm}
\centering
\begin{subfigure}{.1\textwidth}

\tikzstyle{dot}=[rectangle, draw, fill=blue!50,
                        minimum height=5pt, minimum width=10pt]
\begin{tikzpicture}[thick,scale=0.4]
        \node[dot](1) at (0,0) {1};
        \node[dot](4)at (2,0) {4};
        \node[dot](7) at (4,0) {7};
        \node[dot](2) at (0,2) {2};
        \node[dot](3) at (0,4) {3};
        \node[dot](5) at (2,2) {5};
        \node[dot](6) at (2,4) {6};
        \node[dot](8) at (4,2) {8};
        \node[dot](9) at (4,4) {9};
        \draw [<->](1.east)  -- (4.west);
        \draw [<->](1.north)  -- (2.south);
        \draw [<->](2.north)  -- (3.south);
        \draw [<->](3.east)  -- (6.west);
        \draw [<->](4.north)  -- (5.south);
        \draw [<->](7.north)  -- (8.south);
        \draw [<->](5.north)  -- (6.south);
        \draw [<->](6.east)  -- (9.west);
        \draw [<->](9.south)  -- (8.north);
        \draw [<->](2.east)  -- (5.west);
        \draw [<->](4.east)  -- (7.west);
        \draw [<->](8.west)  -- (5.east); 
\end{tikzpicture}
\caption{}
\label{fig:example}
\end{subfigure}%
\begin{subfigure}{.45\textwidth}
\centering
\includegraphics[scale = 0.145]{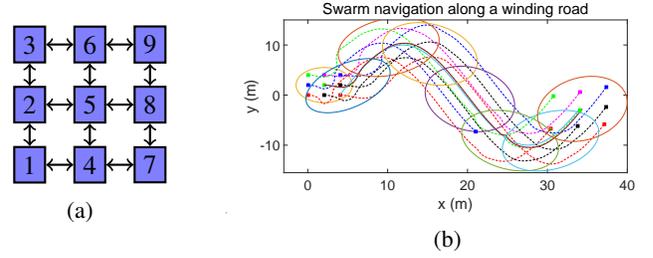}
\caption{}
\label{fig:tratr}
\end{subfigure}%
\caption{ (a) The communication graph of the swarm of robots in the conducted simulation.
(b) Navigation of swarm of robots along a desired trajectory $\zeta$. The robots are initially in a square formation. Their formation evolves to rectangular and parallelogram shapes along the road while tracking $\zeta$.}
\end{figure}

In this Section, we present numerical simulation results to illustrate the performance of our leaderless swarm formation control system. We consider a group of 9 identical rear wheel driving car-like robots with the virtual reference point of each robot located at $D$ $=$ $0.05m$ away from its  center.
 The robots' initial locations, heading angles, and steering angles are given as (0,0,0,0),  (0,2,0,0),  (0,4,0,0),  (2,0,0,0),  (2,2,0,0),  (2,4,0,0),  (4,0,0,0),  (4,2,0,0), and  (4,4,0,0) for Robots 1-9, respectively.  
 The underlying communication graph is given  in Figure \ref{fig:example}. The initial formation is a $4m \times 4m$ square grid, circumscribed by a circle with radius of $3.6091m$.  

\begin{figure}[!h]
\vspace{0.275cm}
\begin{subfigure}{4.3cm}
\includegraphics[scale = 0.11]{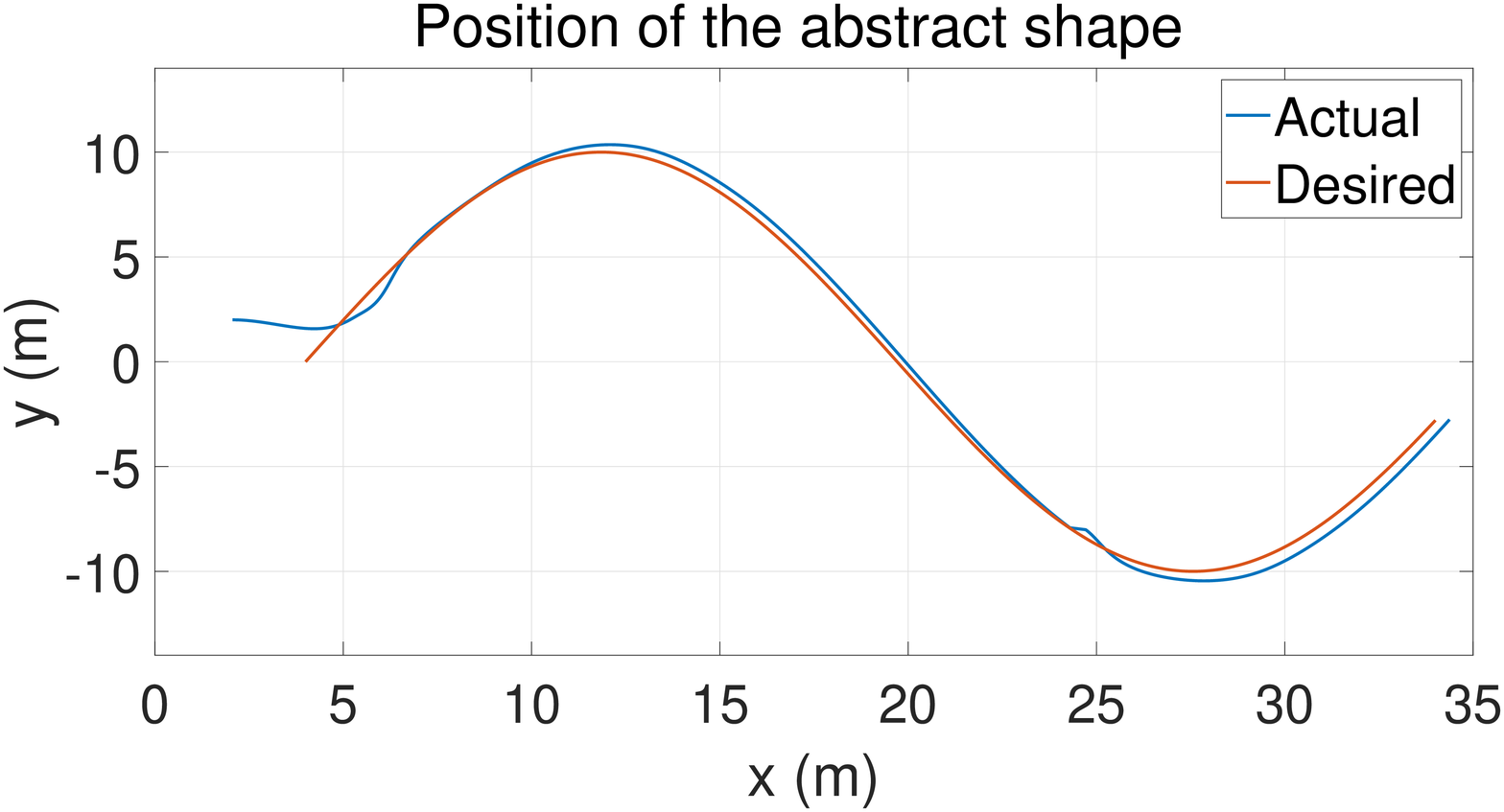}
\caption{}
\label{fig:orient}
\end{subfigure}%
\begin{subfigure}{4.3cm}
\includegraphics[scale = 0.115]{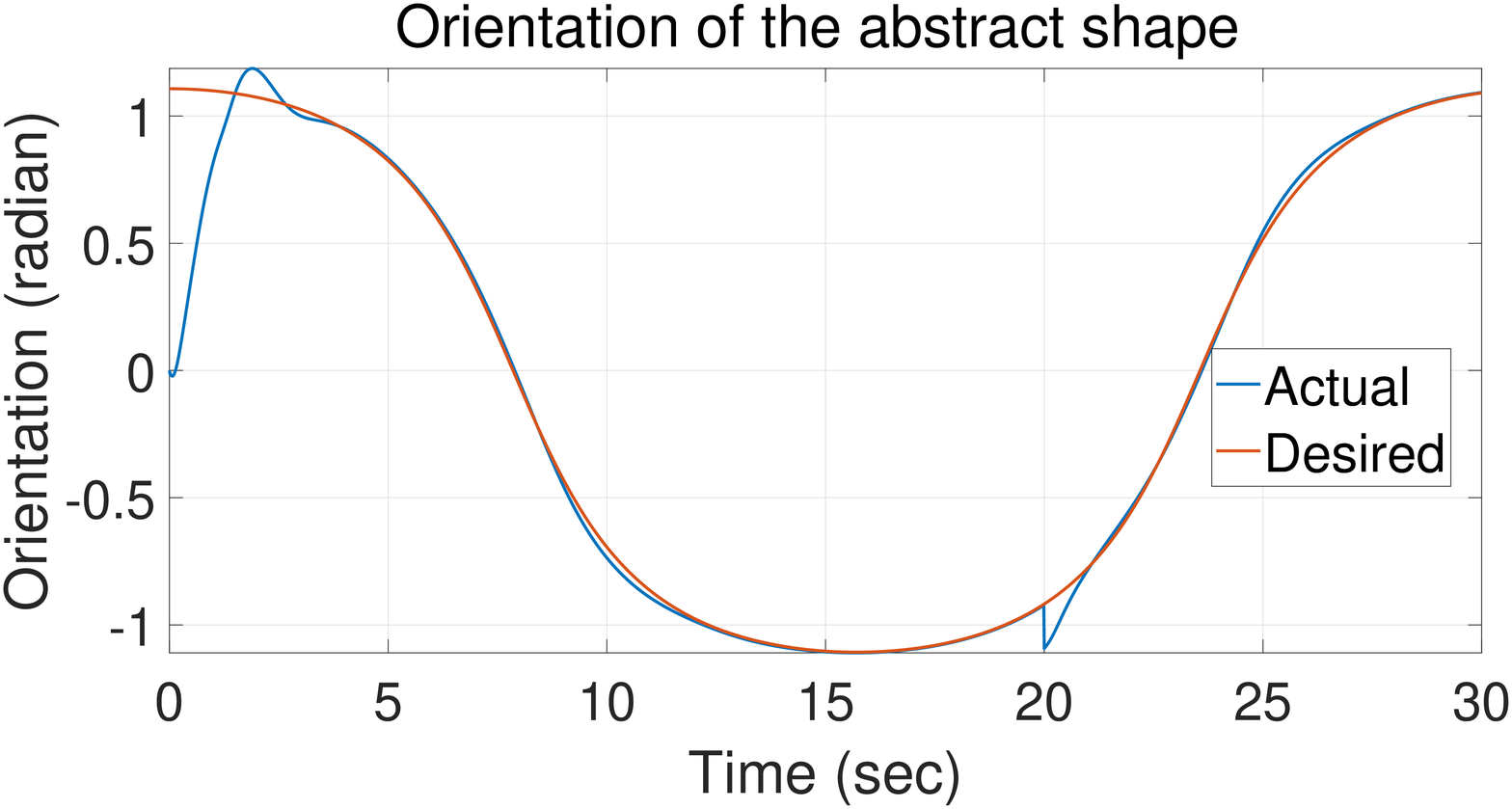}
\caption{}
\label{fig:shape2}
\end{subfigure}\\
\begin{subfigure}{4.3cm}
\includegraphics[scale = 0.11]{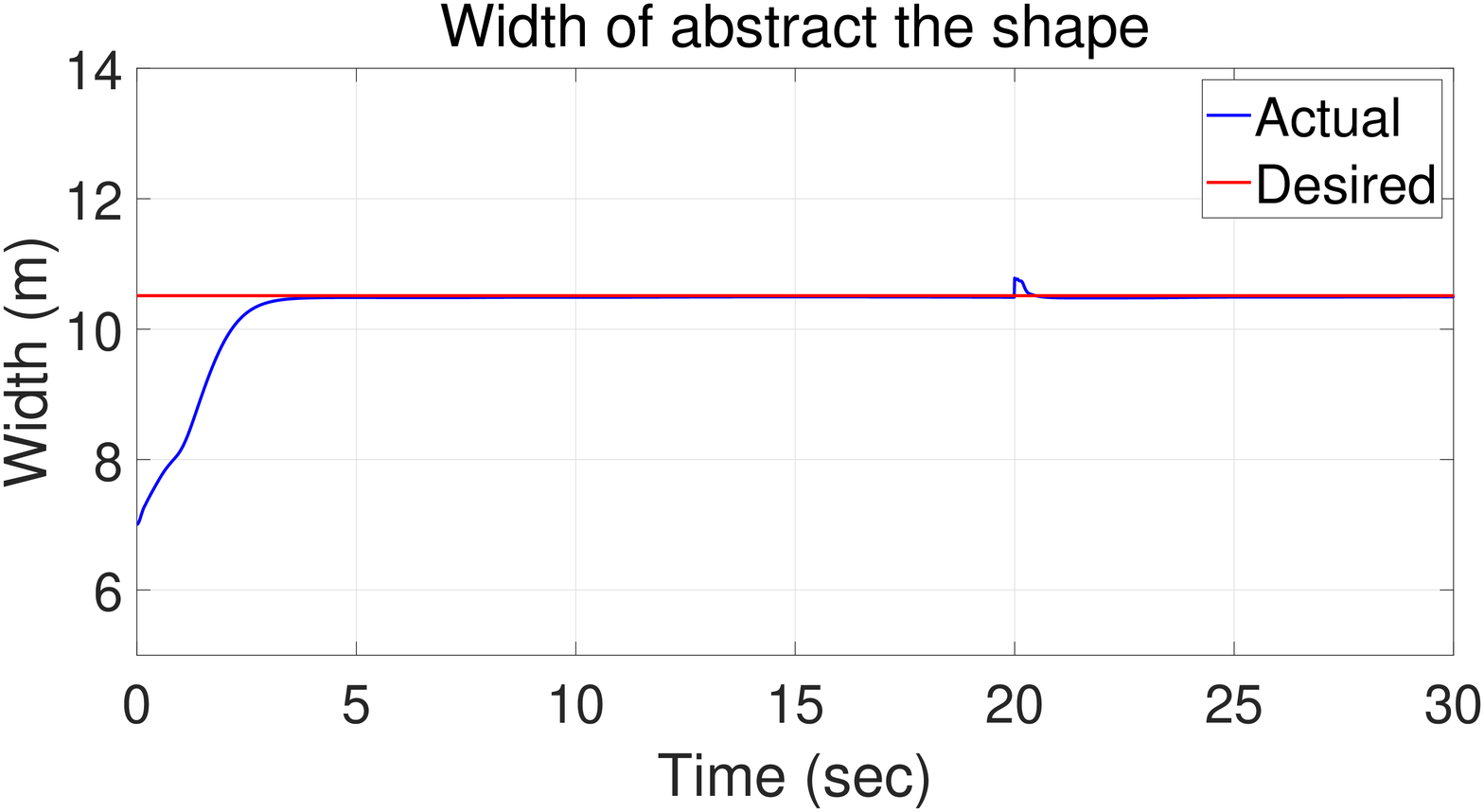}
\caption{}
\label{fig:orient1}
\end{subfigure}%
\begin{subfigure}{4.3cm}
\includegraphics[scale = 0.115]{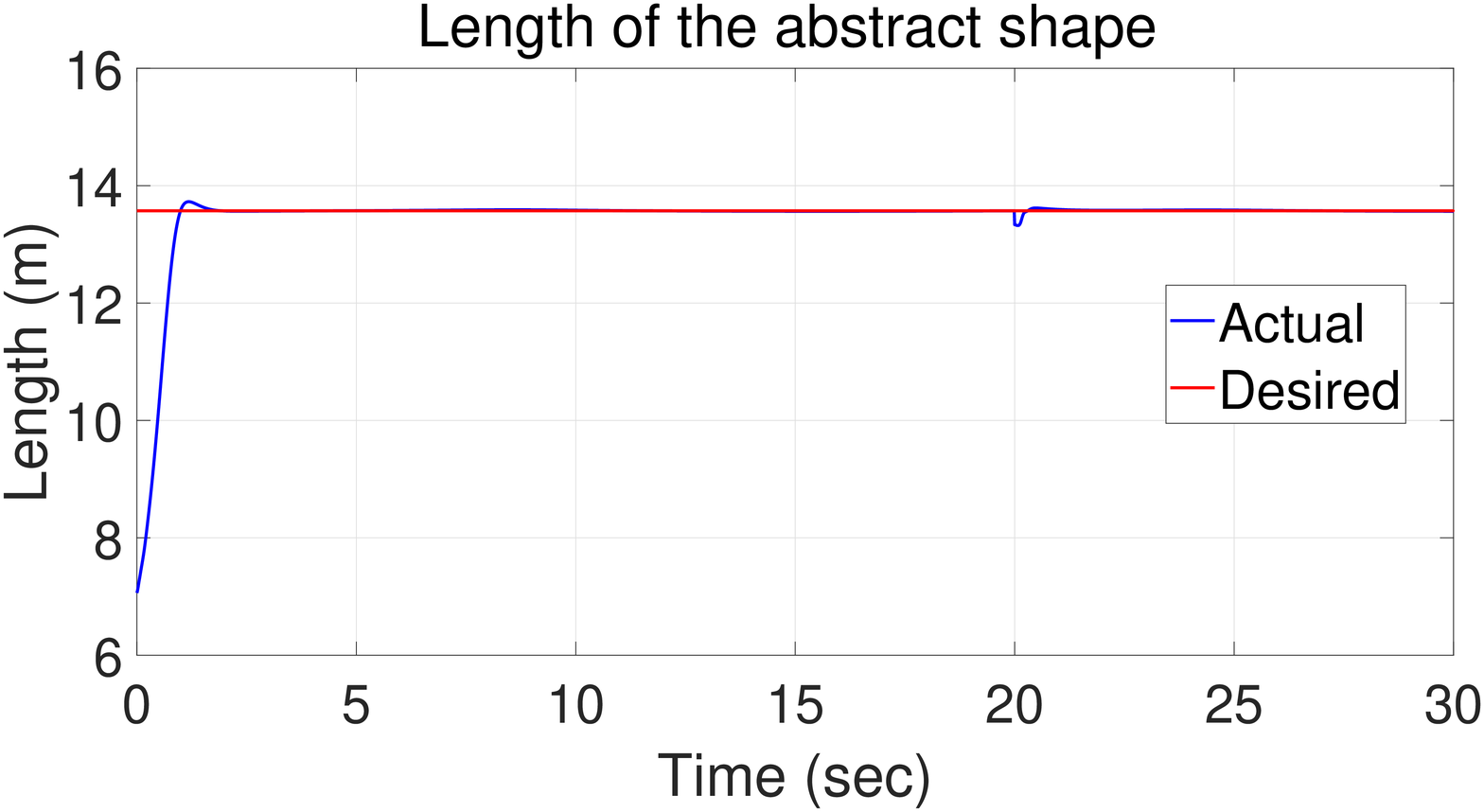}
\caption{}
\label{fig:shape21}
\end{subfigure}%
\caption {The abstract shape parameters trajectory: (a) Swarm position tracking, (b) Swarm orientation tracking, (c) swarm  width tracking, (d) swarm length tracking.}
\label{fig:abstract-state}
\end{figure}
\begin{figure}[!h]
\begin{subfigure}{.25\textwidth}
\includegraphics[scale = 0.115]{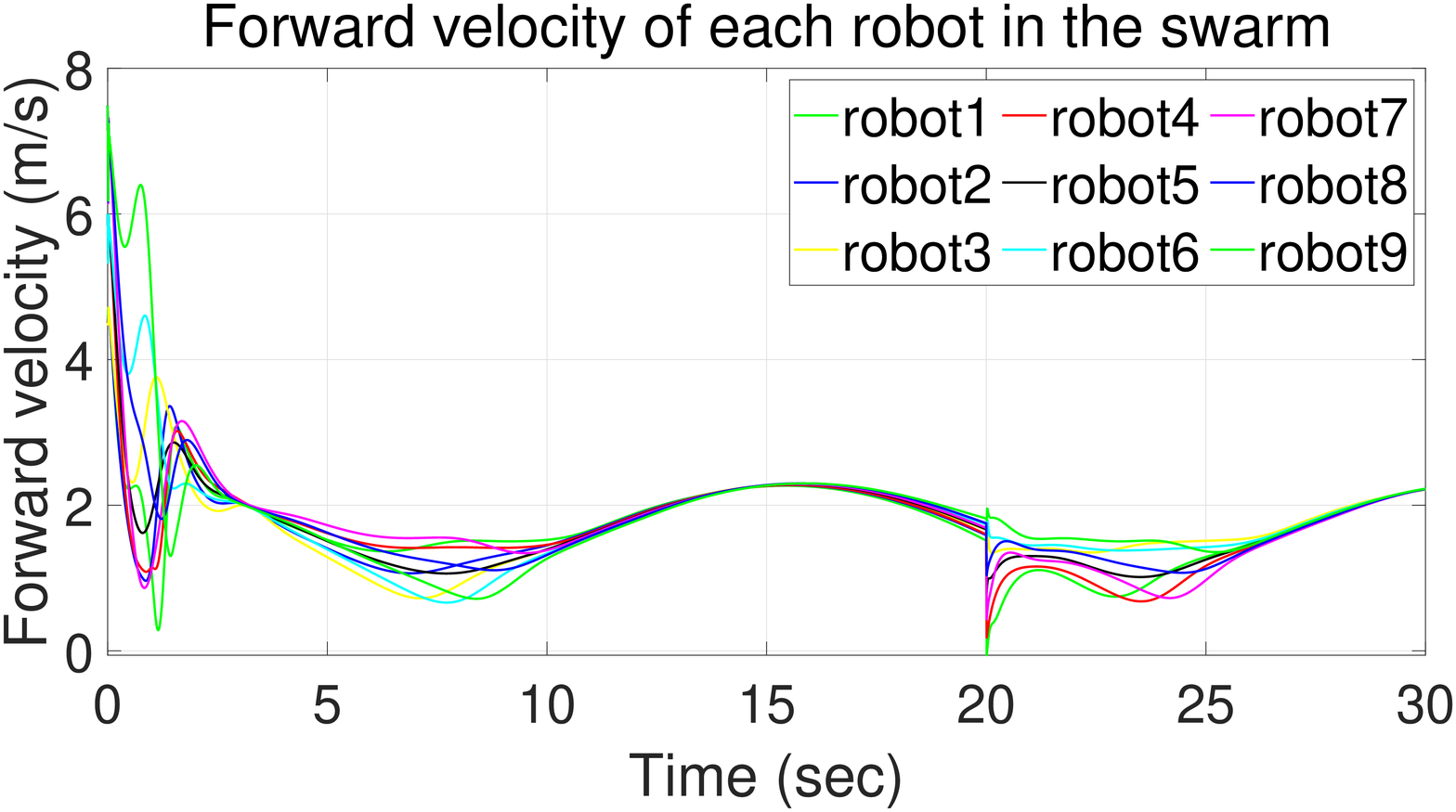}
\caption{}
\label{fig:dvelo}
\end{subfigure}%
\begin{subfigure}{.3\textwidth}
\includegraphics[scale = 0.113]{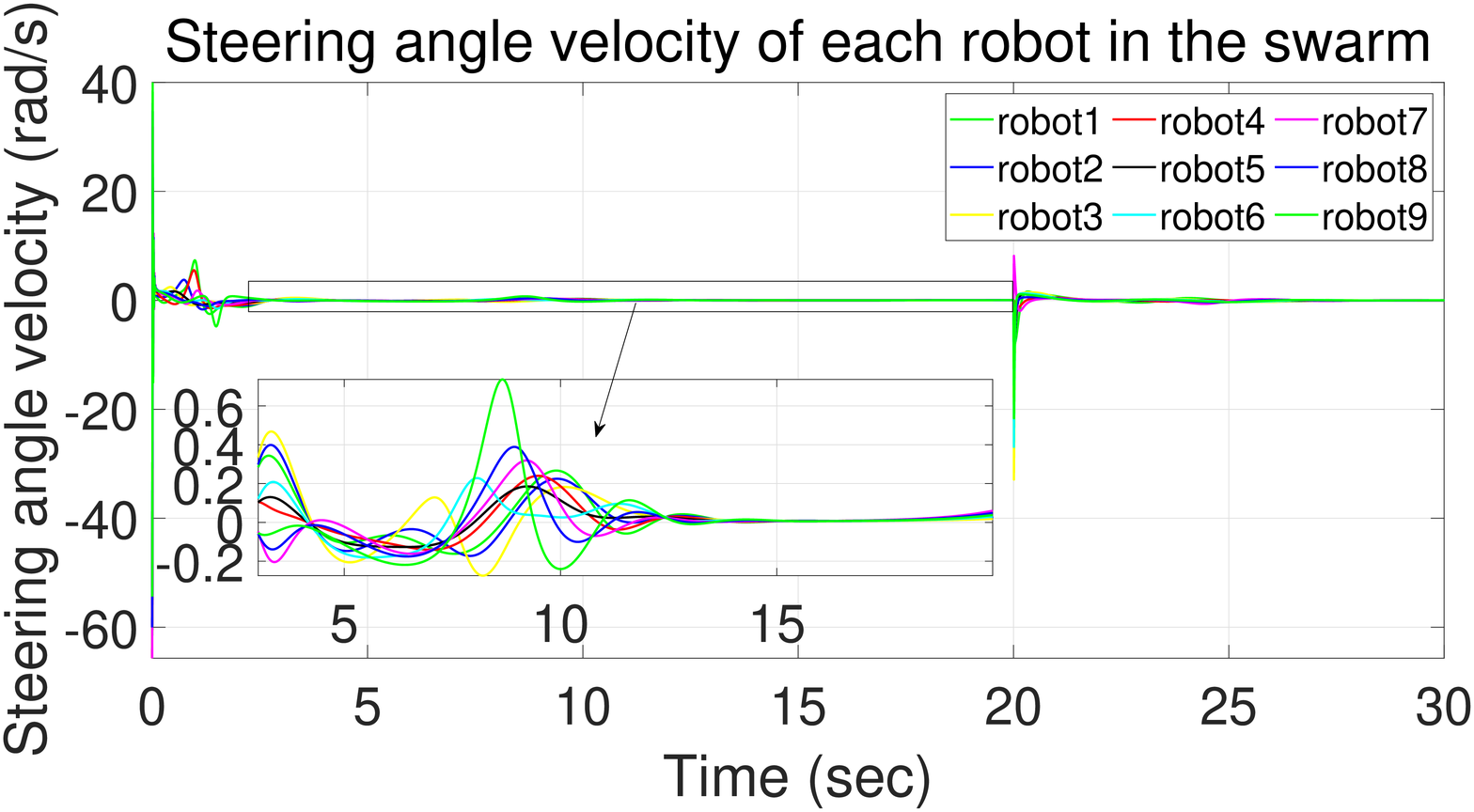}
\caption{}
\label{fig:stevelo}
\end{subfigure}%
\caption{Inputs to the robots: (a) the driving velocity of each robot, (b) the steering angle velocity of each robot.}
\label{fig:inputs}
\end{figure}
\begin{figure}[!h]
\begin{subfigure}{.24\textwidth}
\includegraphics[scale = 0.115]{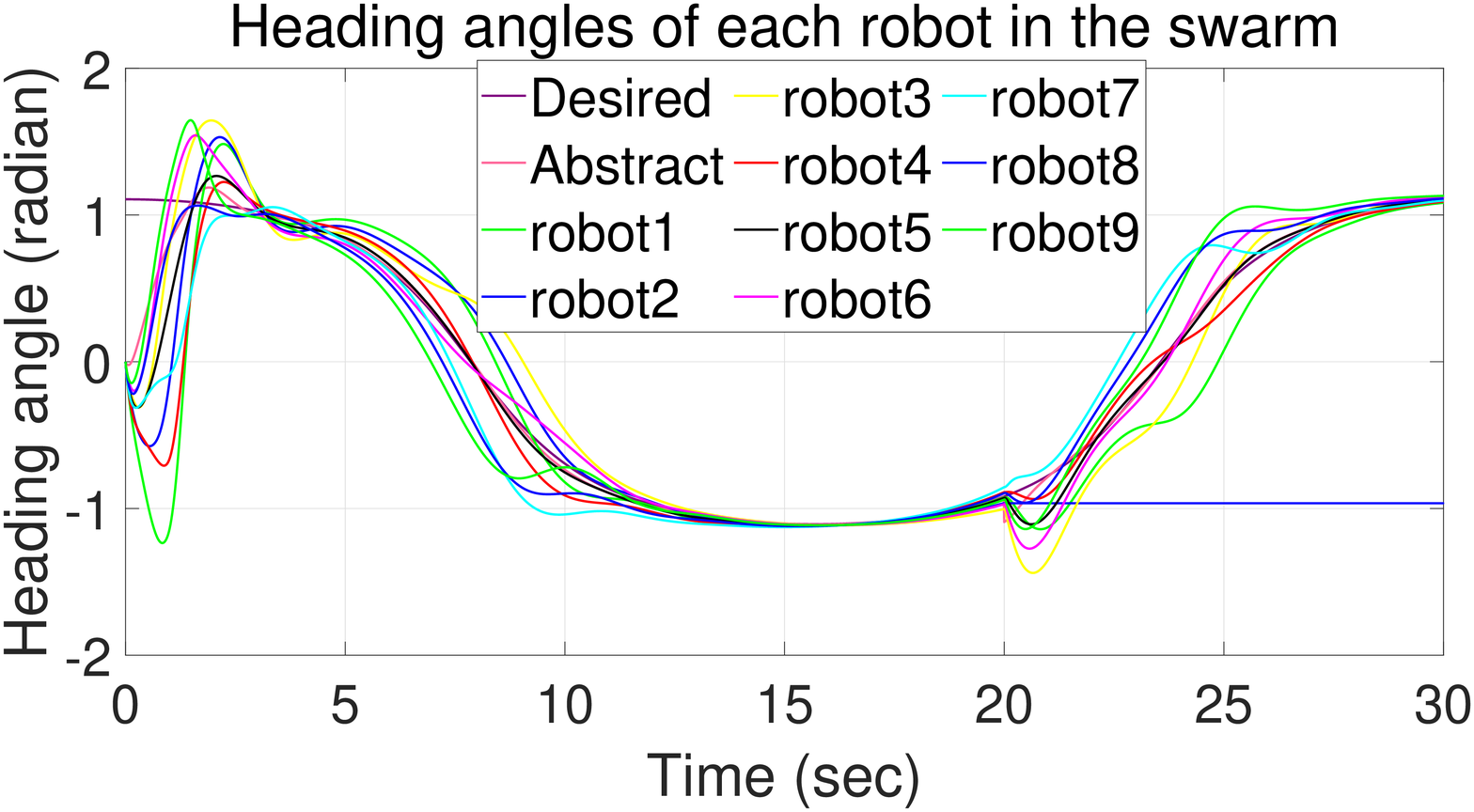}
\caption{}
\label{fig:headn}
\end{subfigure}%
\begin{subfigure}{.2\textwidth}
\includegraphics[width=1.8in,height=0.9in]{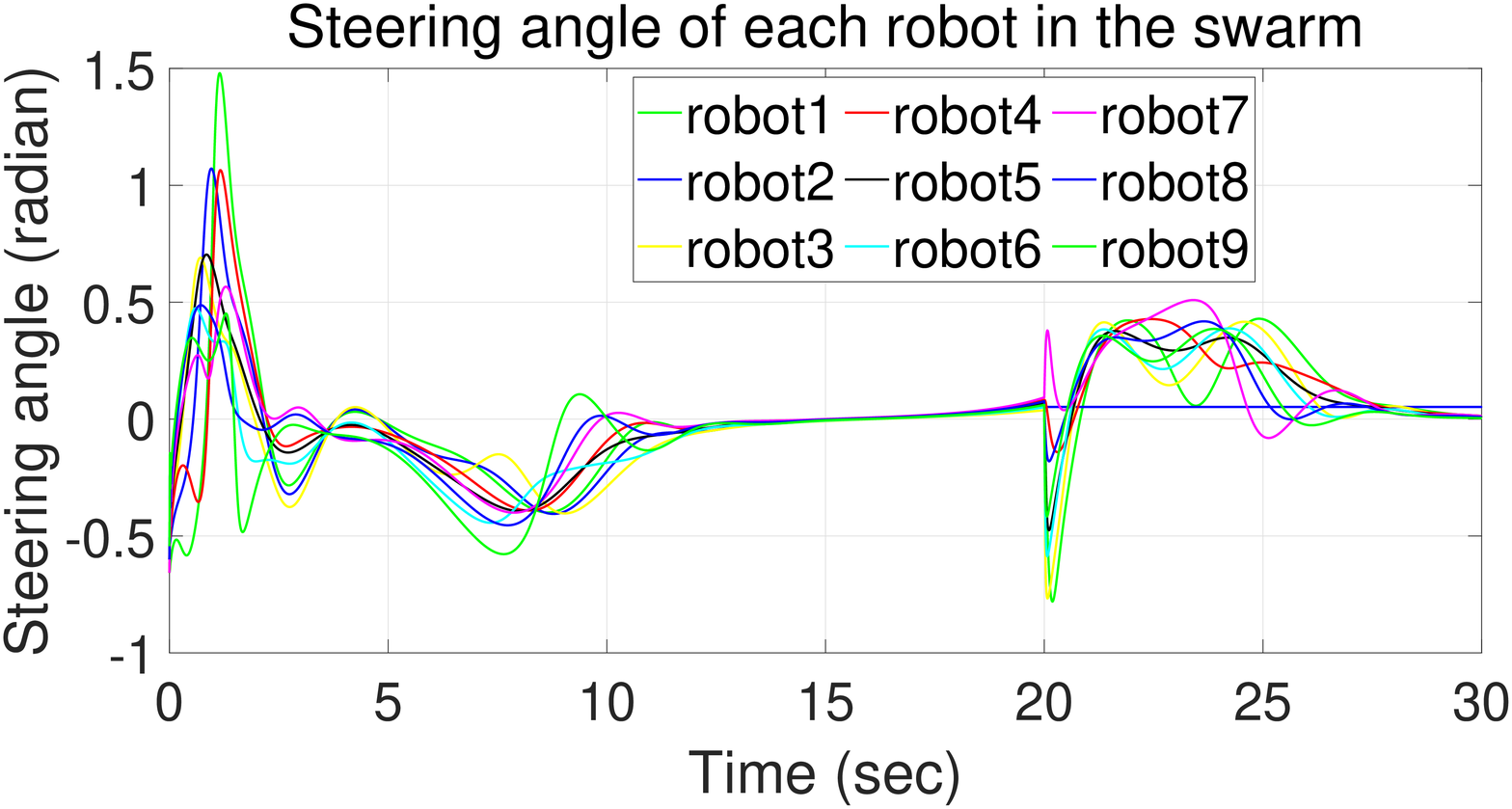}
\caption{}
\label{fig:steerin}
\end{subfigure}%
\caption{(a) The heading angle of each robot, (b) the steering angle of each robot in the swarm.}
\label{fig:angles}
\end{figure}

Consider that the swarm of robots is tasked to navigate along a winding road, given by the trajectory $\zeta (t) = \begin{bmatrix} [\mu_{x_d} & \mu_{y_d}] & \theta_d & s_{2_d} & s_{1_d}\end{bmatrix}= \begin{bmatrix} t+4 &10\sin(0.2t) & \tan^{-1}(2\cos(0.2t))&  10.513& 13.57\end{bmatrix}^\mathrm{T}$ and $\dot{\zeta}(t) = \begin{bmatrix}1& 2\cos(0.2t)&  -0.4\frac{\sin(0.2*t)}{(4*\cos^2(0.2*t)+1)} &0 &0\end{bmatrix}^\mathrm{T}$. The width of the road is  $10.513m$ and we want the length of swarm to be $13.57m$. All the robots have the knowledge of $\zeta$ and $\dot{\zeta}(t)$. After $20s$, we disable Robot 2 to test the robustness of the proposed control algorithm. To track the given path, we use control gains $\bar{K}_i = \begin{bmatrix} 2.5I_{2\times2} & 3 & 0.06 & 0.08 \end{bmatrix}^T$ and $\check{K}_i = \begin{bmatrix} 0.0008 & 0.0008 \end{bmatrix}^\mathrm{T}$ for  the swarm controller and the robot position controller on each Robot $i$, respectively.
The dynamic consensus estimator is used to estimate the values of the abstract shape parameters $a$ to be used in the desired trajectory generation. The consensus parameters are selected to be $\rho = 79$ and $c = 2$. 
Based on this, the simulation results of leaderless swarm formation control are presented in Figure \ref{fig:abstract-state}. 

Looking at individual robots  in Figure \ref{fig:tratr}, the swarm of robots tracks  affine transformations of the initial square grid formation, where the swarm is expanded and elongated by changing its formation between a rectangular and parallelogram shape. 
Figure \ref{fig:tratr} also shows the motion of the abstract shape of the swarm captured by different snapshots of ellipses ($m_a =2,n_a =2$). To further investigate the history of the swarm's configuration, the abstract shape parameters trajectory including $\mu$,  $\theta$,  $s_w$ and $s_l$, are shown in Figure \ref{fig:abstract-state}.  The simulation results in Figure \ref{fig:inputs} show the steering and forward velocity as inputs to individual robots. Further, Figure \ref{fig:angles} shows the heading angle and steering angle of individual robots in the swarm. From these simulation results, it can be observed that the robots in the swarm 
have almost similar velocity and heading angle while navigating the road. Also, the proposed algorithm performs well against robot failures and communication link failures as long as the communication graph, $\mathcal{G}(t)$, remains connected aftermath of the failures.
To demonstrate this, we made $Robot$ $2$ to stop moving at $t=20s$ and disabled its communication links with its neighbours. Accordingly, the simulation results show that after failure of Robot $2$, the swarm again converges to the desired shape and continues tracking the desired trajectory.

\section{conclusion}
\label{sec:form_and_tra_conc}
In this paper, we introduced a distributed swarm formation control framework for transferring a swarm of robots from a current location to the desired location while allowing the shrinkage, expansion, elongation, and compression of the swarm along a reference time-varying path. For this purpose, we represented the swarm by an abstract shape that circumscribes the convex hull of robots' positions. Then, for each robot in the swarm, we designed a distributed control law to track a suitable trajectory that allows the swarm to follow a desired time-varying swarm formation without relying on any leader. We also developed a dynamic average consensus estimator algorithm to estimate the abstract shape states in a distributed manner for use in a trajectory generation. We demonstrated the effectiveness and robustness of the designed control system through simulations by introducing failures to individual robots and their communication links. 

\label{sec:concu}

\section*{Acknowledgment}
This research is supported by Air Force Research Laboratory and OSD under agreement number FA8750-15-2-0116 as well as the National Science Foundation under award number 1832110.
\bibliographystyle{IEEEtran}
\bibliography{ifacconf}
\end{document}